\documentclass[letterpaper, 10 pt, conference]{ieeeconf}  

\IEEEoverridecommandlockouts                              

\usepackage{hyperref}
\usepackage{times}  
\usepackage{helvet}  
\usepackage{courier}  
\usepackage{graphicx} 
\usepackage{cite}  
\usepackage{caption} 
\usepackage{algorithm}
\usepackage{algorithmic}
\usepackage{amsmath} 
\usepackage{xcolor}
\usepackage{colortbl}
\usepackage{dsfont}

\usepackage{subcaption}

\usepackage{caption}
\usepackage{tabularx}
\usepackage{booktabs}
\usepackage{pifont}
\usepackage{amssymb}
\usepackage{multirow}
\usepackage{newfloat}
\usepackage{listings}
\title{\LARGE \bf
Long-Horizon Visual Imitation Learning  via Plan and Code Reflection
}

\author{Quan Chen$^{1,2,*}$,
	Chenrui Shi$^{1,*}$,
	Qi Chen$^{1,2}$,
	Yuwei Wu$^{1}$,
	Zhi Gao$^{1}$,\\
	Xintong Zhang$^{1}$,
	Rui Gao$^{1,2}$,
	Kun Wu$^{3}$,
	Yunde Jia$^{2}$%
	\thanks{$^{*}$ Equal contribution.}
	\thanks{$^{1}$ Beijing Key Laboratory of Intelligent Information Technology, School of Computer Science and Technology, Beijing Institute of Technology, China}
	\thanks{$^{2}$ Guangdong Laboratory of Machine Perception and Intelligent Computing, Shenzhen MSU-BIT University, China}
	\thanks{$^{3}$ Beijing Innovation Center of Humanoid Robotics, China}
	\thanks{Project Website: \url{https://longvil-agent.github.io/}}
}

\begin{document}

\maketitle
\thispagestyle{empty}
\pagestyle{empty}

\begin{abstract}
Learning from long-horizon demonstrations with complex action sequences presents significant challenges for visual imitation learning, particularly in understanding temporal relationships of actions and spatial relationships between objects. In this paper, we propose a new agent framework that incorporates two dedicated reflection modules to enhance both plan and code generation. The plan generation module produces an initial action sequence, which is then verified by the plan reflection module to ensure temporal coherence and spatial alignment with the demonstration video. The code generation module translates the plan into executable code, while the code reflection module verifies and refines the generated code to ensure correctness and consistency with the generated plan. These two reflection modules jointly enable the agent to detect and correct errors in both the plan generation and code generation, improving performance in tasks with intricate temporal and spatial dependencies. To support systematic evaluation, we introduce LongVILBench, a benchmark comprising 300 human demonstrations with action sequences of up to 18 steps. LongVILBench emphasizes temporal and spatial complexity across multiple task types. Experimental results demonstrate that existing methods perform poorly on this benchmark, whereas our new framework establishes a strong baseline for long-horizon visual imitation learning.

\end{abstract}


\section{Introduction}

Visual imitation learning (VIL) aims to imitate human actions by observing demonstration videos, without relying on explicit supervision or environment interaction~\cite{firstVIL,firstVIL2,li2024robust}. 
By bridging high-level video understanding and low-level robotic control, VIL offers a general framework for teaching models to perform tasks in a wide range of physical and simulated settings, such as object manipulation in robotics~\cite{VIL3npis,VIL2025,jonnavittula2025view}, solely from human demonstration videos.

Recent advances have integrated vision-language models (VLMs) into VIL, improving the generation of symbolic representations or executable control codes from video~\cite{mu2023embodiedgpt,patel2023pretrained,wang2023demo2code}. These models have demonstrated strong performance on short-horizon tasks involving simple, atomic actions, typically 1–5 steps, such as, pressing a button or moving an object. In such cases, temporal-spatial dependencies are minimal, and errors are unlikely to accumulate.

However, 
many real-world tasks are long-horizon, comprising both complex temporal action sequences and diverse spatial relationships~\cite{gupta2019relaypolicylearningsolving,yang2025lohovla,grunde2021agqa}. 
For instance, building a block tower requires placing components in a strict order, where early mistakes can make later steps infeasible~\cite{pirk2020modelinglonghorizontaskssequential}.
Current VLM-based approaches struggle in these scenarios, often producing misaligned plans, wrong spatial relationships and even hallucinated actions, due to limited temporal-spatial reasoning of VLMs and the lack of mechanisms for error detection and correction.


In this paper, we propose a new agent framework that combines plan and code generation with two reflection modules: plan reflection and code reflection. 
The plan generation module produces an initial action sequence. Then, the plan reflection module verifies whether each inferred action in the sequence aligns with the observed visual content, including both the temporal ordering of these actions and spatial relationships between objects.
The code generation module translates the plan into executable code, while the code reflection module ensures that the generated code logically matches the action sequence, allowing correction before execution. 
These two reflection mechanisms transform imitation into a cycle of planning, verification, and correction, significantly improving the robustness and reducing error propagation in long-horizon tasks. 


To support evaluation, we introduce LongVILBench, a new benchmark specifically designed to assess visual imitation learning under long-horizon tasks. Unlike existing benchmarks that focus on short demonstrations with 1-8 atomic actions~\cite{imitrob,fetchbench,seedo}, LongVILBench consists of 300 human demonstration videos covering 150 tasks, containing 1 to 18 atomic actions in three levels of complexity.
These tasks are recorded in different visual conditions, spanning four types of manipulation domains and annotated with structured plans and code.
LongVILBench enables systematic evaluation of an agent’s capabilities across perception, reasoning, planning, and execution for long-horizon visual imitation learning. 
We validate our framework on LongVILBench through extensive experiments across multiple baselines and metrics. Results show that our method handles long-horizon demonstrations more effectively than existing approaches, highlighting the benefits of incorporating reflection into long-horizon visual imitation learning.

\begin{figure*}[t]
    \centering
    \includegraphics[width=0.8\linewidth]{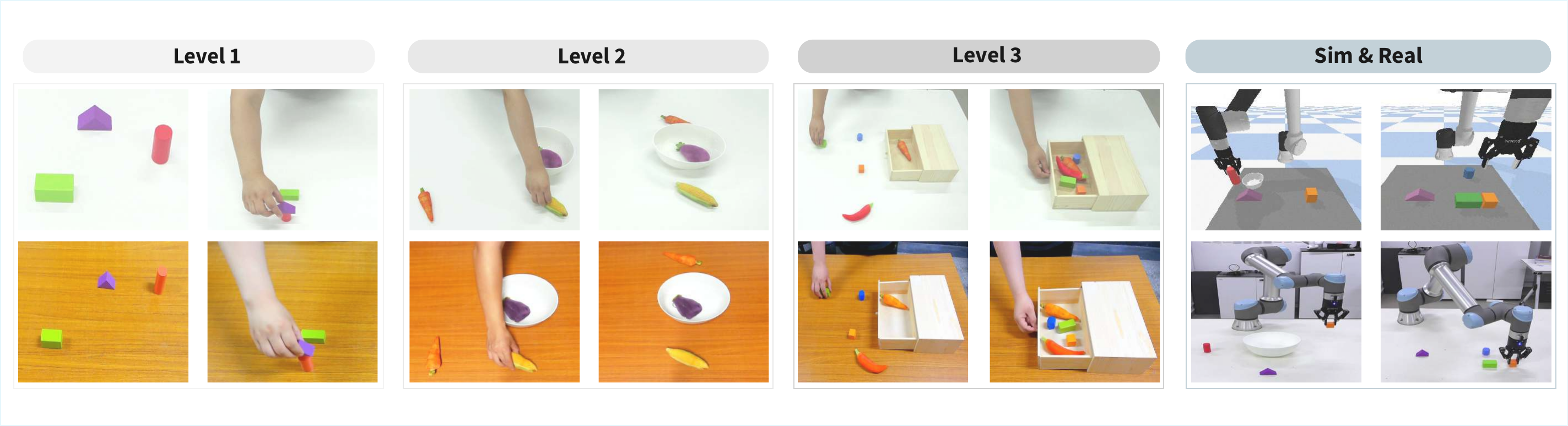}
    \caption{Example images from LongVILBench: LongVILBench includes 150 tasks and 300 human demonstration videos, with tasks grouped into three levels based on the number of actions involved. The generated codes can be verified both in real world and simulation.}
    \label{fig:dataset_overview}
\end{figure*}

Our contributions are as follows:

\begin{itemize}
    
    \item We propose a VLM-based agent framework with two reflection modules that verify and refine the generated plans and codes.
    \item We introduce LongVILBench, a new benchmark with 300 human demonstration videos covering 150 tasks of varying complexity, specifically designed to evaluate long-horizon visual imitation learning models. Figure~\ref{fig:dataset_overview} illustrates representative tasks sampled from LongVIL-Bench, covering different manipulation domains and complexity levels.
    \item We conduct experiments across multiple baselines and settings, demonstrating superior performance of our framework. Our results highlight the importance of two reflection modules in long-horizon visual imitation learning.
\end{itemize}

\section{Related Works}
\subsection{VLMs for Visual Imitation Learning}
Vision-language models (VLMs) have recently been adopted to enhance visual imitation learning (VIL), particularly in high-level planning and low-level control. For planning, VLMs map demonstrations to symbolic action sequences or code, enabling subtask decomposition and policy synthesis~\cite{xie2025roboticprogrammervideoinstructed,liang2022code,wake2024gpt,seedo,driess2023palmeembodiedmultimodallanguage,wei2022chain}. For control, VLMs bypass handcrafted primitives by directly grounding actions from visual inputs~\cite{chen2024vlmimic,huang2023voxposer,zhao2025manipbench,smith2024steer,zitkovich2023rt,team2024octo,kim2024openvla}. Despite these advances, current VLM-based VIL methods are mostly limited to short-horizon, simple tasks, due to: (1) limited context length, restricting the ability to model long demonstrations; and (2) over-reliance on VLMs' predictions and lack of self-correction. These limitations hinder generalization to long, compositional tasks with complex spatial-temporal dependencies. Different from existing methods, we propose a new agent framework with two reflection modules for self-verification at both plan and code levels. 




\subsection{Benchmarks for Visual Imitation Learning}

Existing benchmarks for VIL have primarily focused on short-horizon tasks with limited structural complexity. As summarized in Table~\ref{tab:dataset_compare}, Imitrob~\cite{imitrob} and FetchBench~\cite{fetchbench} consist of extremely short tasks (1–5 steps), offering minimal temporal dependencies and limited spatial reasoning. SeeDo~\cite{seedo} incorporates temporal sequences and spatial relations, yet its spatial interactions are limited in diversity, with most tasks involving simple object placements in constrained configurations. In contrast, our proposed benchmark LongVILBench supports long-horizon tasks (1–18 steps) with rich temporal dependencies, six types of spatial relations, and three levels of hierarchical difficulty. This enables fine-grained evaluation of a model’s temporal planning, spatial grounding, and robustness across compositional and long-horizon scenarios.

\subsection{Self-Reflection and Verification Mechanisms}

Recent studies have introduced reflection and verification mechanisms to improve reasoning and error correction in complex tasks. These methods typically introduce feedback loops to refine plans or actions through iterative evaluation.
Most existing approaches adopt a text-driven reflection paradigm~\cite{madaan2023self,shinn2023reflexion,yu2024exact,yuan2025remac,Gao_2024_CVPR}, where large language models (LLMs) revise task plans or code based on execution results or predicted states. These systems rely on natural language instructions and perform reflection by checking whether the task outcomes align with textual goals or preconditions.
A smaller body of work explores multimodal or video-based reflection~\cite{cheng2024vision,feng2025reflective,meng2025data,li2025iterative,gaomulti}, where models incorporate visual and temporal information to detect inconsistencies and guide corrections. This enables more grounded and fine-grained reasoning over spatial configurations and action sequences.
In contrast to prior work, our method introduces dual reflection modules that operate on both video observations and generated code. By reasoning over spatial-temporal dependencies in demonstration videos and verifying semantic consistency in code, our approach enables more reliable long-horizon imitation.



\begin{table}[t]
\centering
\small                 
\setlength{\tabcolsep}{2.5pt} 
\begin{tabular*}{\linewidth}{l@{\extracolsep{\fill}}|c|c|c|c|c}
\toprule
\textbf{Benchmark} & \textbf{Data} & \textbf{Length} & \textbf{Spatial} & \textbf{Tasks} & \textbf{Diff.} \\
\midrule
Imitrob & videos  & 1 & 0 & 56 & \ding{55} \\
FetchBench & images  & 3--5 & 5 & 6156 & \ding{55} \\
SeeDo & videos  & 2--8 & 4 & 106 & \ding{51} \\
\textbf{Ours} & videos & 1--18 & 6 & 150 & \ding{51} \\
\bottomrule
\end{tabular*}
\caption{Comparison of our benchmark with existing benchmarks in terms of data types (Data), task length (Length), spatial relationship types (Spatial), task numbers (Tasks), difficulty levels (Diff.). }
\label{tab:dataset_compare}
\vspace{-0.7cm}    
\end{table}
\section{Method}
\subsection{Overview}
\begin{figure*}
    \centering
    \includegraphics[width=0.8\linewidth]{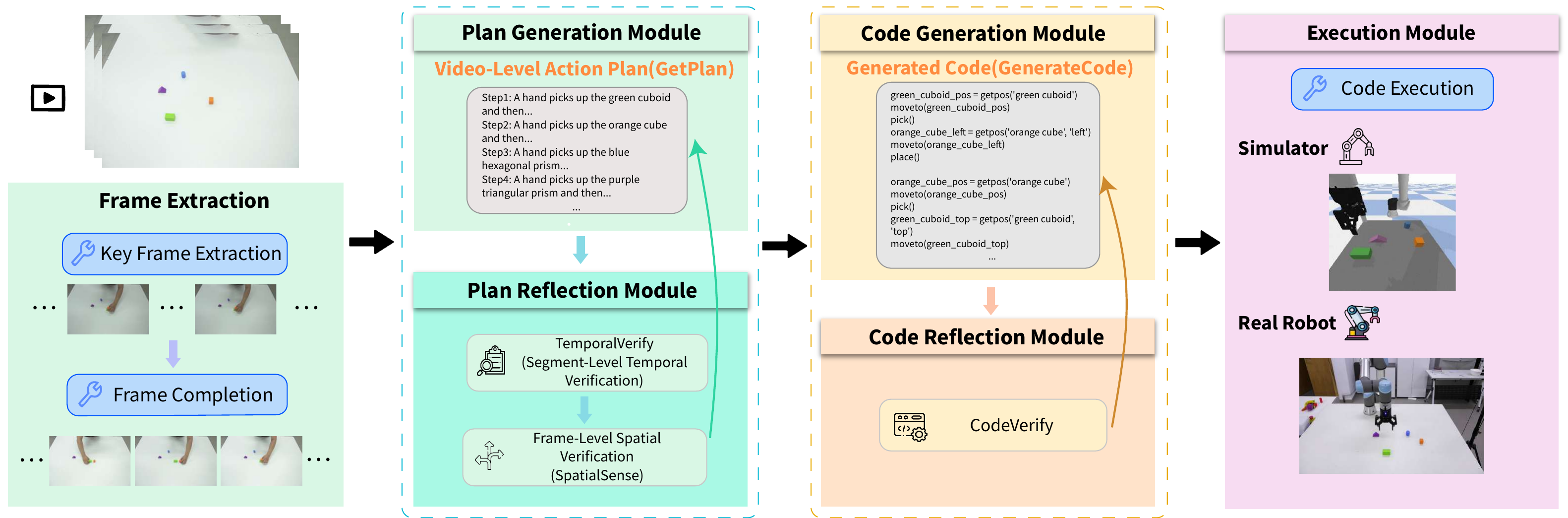}
    \caption{ An overview of our agent framework. Our agent framework is comprised of four key modules: plan generation module, plan reflection module, code generation module, and code reflection module. Together, our agent framework turns a human demonstration video into a code program, which can be executed in a simulator or a real-world robot. }
    \label{fig:method_overview}
\end{figure*}

We present \textbf{LongVIL}, an agent framework designed for long-horizon visual imitation learning. 
The input of our agent framework is a human demonstration video, and the output is a code program describing the action sequence of the demonstration video. 
Our agent is comprised of a plan generation module, a plan reflection module, a code generation module and a code reflection module. 
Specifically, the overall architecture is depicted in Figure~\ref{fig:method_overview}, 
(1) a plan generation module $\mathcal{G}_{\rm plan}$ that extracts temporally aligned key frames and generates an initial action plan; 
(2) a plan reflection module $\mathcal{R}_{\rm plan}$ that verifies and refines this plan for temporal and spatial consistency with the video; 
(3) a code generation module $\mathcal{G}_{\rm code}$ that synthesizes executable robot code from the validated plan; and
(4) a code reflection module $\mathcal{R}_{\rm code}$ that ensures alignment between code and the validated plan. The process can be expressed as
\begin{equation}
    \Pi = \mathcal{R}_{\rm code} \big( \mathcal{G}_{\rm code} \big( \mathcal{R}_{\rm plan} \big( \mathcal{G}_{\rm plan}(V) \big) \big) \big),
\end{equation}
where $V$ denotes the input human demonstration video and $\Pi$ is the final verified executable code program for manipulation.

\subsection{Plan Generation Module ($\mathcal{G}_{\rm plan}$)}
The plan generation module $\mathcal{G}_{\rm plan}$ tasks the human demonstration video $V = \{f_t\}_{t=1}^{T'}$ as input, where $f_t$ denotes the $t$-th video frame, and constructs an action plan $\mathcal{A}$ that encodes temporal ordering of these actions and spatial relationship of multiple objects. 
The generation of plans follows the following three processes: \textit{Keyframe Extraction}, \textit{Keyframe Completion}, and \textit{Video-Level Action Plan generation}.


\subsubsection{Keyframe Extraction}
For each frame $f_t$, we extract the 3D hand position $h_t \in \mathbb{R}^3$ using the hand pose estimation model MediaPipe~\cite{lugaresi2019mediapipe}. We then compute the instantaneous hand velocity $v_t$ as
\begin{equation}
    v_t = \| h_{t+1} - h_t \|_2.
\end{equation}
Keyframes $\mathcal{K} \subset V$ are selected based on local minima in the smoothed velocity profile,
\begin{equation}
    \mathcal{K} = \left\{ f_t \;\middle|\; v_t < v_{t-1},\ v_t < v_{t+1} \right\}.
\end{equation}
This process identifies temporally informative frames, since velocity local minima typically indicates key action manipulation moments such as picking or placing objects~\cite{seedo,yanokura2020understanding}.




\subsubsection{Keyframe Completion}
To capture rapid or transient actions potentially missed by sparse velocity minima, we augment the initial keyframe set by inserting additional frames between distant keyframes.

Let $\mathcal{F}_{\text{hand}} \subseteq V$ denote the set of frames with detected hands. For each adjacent pair $f_i, f_j \in \mathcal{K}$ satisfying $|j - i| > \Delta$, we compute two candidate indices at fractional positions:
\[
k_1 = \left\lfloor i + \frac{1}{3}(j - i) \right\rfloor,\quad
k_2 = \left\lfloor i + \frac{2}{3}(j - i) \right\rfloor.
\]
Compared to midpoint sampling, this two-point scheme better recovers consecutive missing actions. We then locate the nearest valid hand frames:
\[
\tilde{k}_1 = \arg\min_{k \in \mathcal{F}_{\text{hand}}} |k - k_1|,\quad
\tilde{k}_2 = \arg\min_{k \in \mathcal{F}_{\text{hand}}} |k - k_2|.
\]
The final keyframe set is
\begin{equation}
    \mathcal{K}^* = \mathcal{K} \cup \{ f_{\tilde{k}_1},\ f_{\tilde{k}_2} \},
\end{equation}
ensuring that only semantically meaningful frames with visible hand motion are inserted.

\subsubsection{Video-Level Action Plan Generation}
Given the keyframe set $\mathcal{K}^*$ and the detected object set $O$, we use a vision-language model (VLM) as \textit{GetPlan} to generate the action plan,
\begin{equation}
    \mathcal{A} = \texttt{GetPlan}(\mathcal{K}^*, O),
\end{equation}
where $\mathcal{A} = [\alpha_1, \dots, \alpha_T]$, and each action $\alpha_i \in \mathcal{A}$ is a tuple:
\begin{equation}
    \alpha_i = \langle a_i,\ S_i,\ e_i \rangle.
\end{equation}
Here, $a_i$ denotes the natural language description of the $i$-th action, $S_i$ is a list of frame indices indicating the temporal span during which the action occurs, and $e_i$ provides a natural language justification explaining why this action is categorized as such.

After the aforementioned three processes, the agent generates the initial action plan, denoted as $\mathcal{A}$.

\subsection{Plan Reflection Module ($\mathcal{R}_{\rm plan}$)}
The plan reflection module, denoted as $\mathcal{R}_{\rm plan}$, refines the initial action plan $\mathcal{A} = [\alpha_1, \dots, \alpha_T]$ by verifying its consistency with the human demonstration video. Specifically, it assesses both the temporal ordering of actions and the spatial relationships between involved objects.
To perform this verification, $\mathcal{R}_{\rm plan}$ automatically invokes two specialized tools: \textit{Segment-Level Temporal Verification}, which ensures the global temporal coherence of the action sequence, and \textit{Frame-Level Spatial Verification}, which checks whether the spatial configurations of objects at the frame level are consistent with the demonstration videos.
If discrepancies are detected, $\mathcal{R}_{\rm plan}$ would use \textit{CorrectPlan} to update $\mathcal{A}$ accordingly to produce a refined and more faithful plan .

\subsubsection{Segment-Level Temporal Verification}
To verify the temporal consistency of each action, the plan reflection module evaluates every action tuple $\alpha_i = \langle a_i, S_i, e_i \rangle \in \mathcal{A}$ using a vision-language model (VLM) as follows:
\begin{equation}
    \langle l_i^{\text{seg}},\ e_i^{\text{seg}} \rangle = \texttt{TemporalVerify}(S_i, a_i),
\end{equation}
where $l_i^{\text{seg}} \in \{\texttt{Yes}, \texttt{No}, \texttt{Unclear}\}$ denotes the verification label indicating whether the action $a_i$ is temporally consistent with the segment $S_i$, and $e_i^{\text{seg}}$ provides a textual explanation justifying the decision based on visual evidence.

\subsubsection{Frame-Level Spatial Verification}
To validate spatial relationships, the agent examines whether the final frame of each segment $S_i$—denoted as $f_i^{\text{end}}$—satisfies the spatial constraints expressed in the action description $a_i$ of that segment. 
This verification is conducted using a vision-language model (VLM), referred to as \texttt{SpatialVerify}:
\begin{equation}
    s_i = \texttt{SpatialVerify}(f_i^{\text{end}}, a_i),
\end{equation}
where $s_i$ denotes the predicted spatial relation inferred from frame $f_i^{\text{end}}$. This predicted relation is then compared against the expected spatial semantics described in $a_i$ to assess consistency.



\subsubsection{Plan Correction}
The corrected action $a_i^*$ is defined as
\begin{equation}
    a_i^* =
    \begin{cases}
        a_i, & \text{if valid} \\
     \texttt{CorrectPlan}(a_i,\ e_i^{\text{seg}},\ s_i),  &\text{otherwise}, 
    \end{cases}
\end{equation}
where \texttt{CorrectPlan} is an VLM, `valid' indicates that $a_i$ passes both segment-level and frame-level verification. 
This revision may involve adjusting segment boundaries, refining object references, or correcting spatial relationships. By applying this correction process to all actions in the initial plan, the agent produces a refined and validated plan $\mathcal{A}^* = [\alpha_1^*, \dots, \alpha_T^*]$, where each $\alpha_i^*$ represents the updated action tuple. 

\subsection{Code Generation Module ($\mathcal{G}_{code}$)}
The code generation module $\mathcal{G}_{code}$ synthesizes an initial executable program $\Pi = [\pi_1, \dots, \pi_T]$ from the refined action plan $\mathcal{A}^* = [a_1^*, \dots, a_T^*]$. The generation is performed by a VLM \texttt{GenerateCode}, which maps each action $a_i^*$ to a code snippet $\pi_i$ based on a predefined set of robot motion primitives, 
\begin{equation}
    \pi_i = \texttt{GenerateCode}(a_i^*),
\end{equation}
where $\pi_i$ consists of predefined function calls of atomic actions. There are a total number of seven function calls, including functions for moving, picking, opening drawer and so on. Please refer to Supplementary Materials for the complete table of these predefined atomic actions.

\subsection{Code Reflection Module ($\mathcal{R}_{\rm code}$)}
The code reflection module $\mathcal{R}_{\rm code}$ verifies the correctness of the initial program $\Pi = [\pi_1, \dots, \pi_T]$, and corrects any mistakes until each generated code block $\pi_i$ aligns with the action $a_i^*$ in $\mathcal{A}^*$. 
Verification and correction are performed by a VLM \texttt{CodeVerify}.


For each action–code pair $(a_i^*, \pi_i)$, the verification function returns
\begin{equation}
    \langle l_i^{\text{code}},\ e_i^{\text{code}} \rangle = \texttt{CodeVerify}(a_i^*, \pi_i),
\end{equation}
where $l_i^{\text{code}} \in \{\texttt{Yes}, \texttt{No}\}$ indicates semantic alignment, and $e_i^{\text{code}}$ provides reasoning to support the decision.


If misalignment is detected, the corrected code $\pi_i^*$ is obtained as
\begin{equation}
    \pi_i^* =
    \begin{cases}
        \pi_i, & \text{if } l_i^{\text{code}} = \texttt{Yes}, \\
        \texttt{CorrectCode}(a_i^*,\ \pi_i,\ e_i^{\text{code}}), & \text{otherwise},
    \end{cases}
\end{equation}
where \texttt{CorrectCode} is an LLM. 
The correction procedure updates spatial arguments, object names, or atomic action types. This reflection process yields the final executable program $\Pi^* = [\pi_1^*, \dots, \pi_T^*]$.

\subsection{Execution}
The execution module $\mathcal{E}$ deploys the verified program $\Pi^*$ either in simulation, \textit{e.g.}, PyBullet~\cite{coumans2016pybullet}, MuJoCo~\cite{todorov2012mujoco}, or on physical robotic platforms, \textit{e.g.}, UR5e~\cite{ur5e}, Franka Emika Panda~\cite{franka}. Each predefined atomic action is mapped to the corresponding low-level motor commands via backend-specific controllers. 

\section{LongVILBench}
Existing benchmarks in VIL primarily focus on short-horizon tasks.
These benchmarks are insufficient for evaluating long-horizon tasks. To support more systematic evaluation, we introduce \textbf{LongVILBench}, a comprehensive benchmark for long-horizon VIL.


\subsection{Formulation}

Each task instance in LongVILBench is modeled as a structured tuple,
\begin{equation}
\mathcal{D} = (\mathcal{O}, \mathcal{P}, \mathcal{V}, \mathcal{A}, \boldsymbol{\Pi}),
\label{eq:dataset_formulation}
\end{equation}
where
\begin{itemize}
    \item $\mathcal{O} = \{o_1, o_2, \dots, o_n\}$ denotes the set of physical objects involved in the task;
    \item $\mathcal{P} = \{\text{pos}(o_1), \dots, \text{pos}(o_n)\}$ represents the initial 3D positions of these objects;
    \item $\mathcal{V} = \{f_1, f_2, \dots, f_{T'}\}$ is the visual demonstration, consisting of $T'$ RGB frames, where $f_t$ denotes the frame at step $t$;
    \item $\mathcal{A} = [a_1, a_2, \dots, a_T]$ is a temporally ordered sequence of $T$ atomic actions, each $a_t$ conforming to a structured action, defined as $a_t=[\alpha,o,d]$, $\alpha$ is the atomic operation, $o$ is the object, $d$ is spatial relation. 
    \item $\boldsymbol{\Pi} = \{\pi_1, \pi_2, \dots, \pi_T\}$ is the symbolic program, where each $\pi_t$ is an executable robot code corresponding to an action.
\end{itemize}

\subsection{Task Category}
 All tasks in LongVILBench can be categorized into three representative manipulation categories: \textit{block manipulation}, \textit{tabletop cleanup}, and \textit{vegetable sorting}. 
All tasks are composed of four atomic operations: \textit{pick}, \textit{place}, \textit{open}, and \textit{close}. Interactions occur between 14 unique objects, and spatial relations are defined using six directional predicates: \textit{left}, \textit{right}, \textit{front}, \textit{behind}, \textit{on top of}, and \textit{into}.
Tasks are categorized into three difficulty levels based on action sequence length: Level 1 Short Tasks (1–4 actions), Level 2 Medium Tasks (5–8 actions), and Level 3 Long Tasks (9–18 actions).

\subsection{Evaluation Metrics}

To comprehensively evaluate model performance, we introduce three evaluation metrics: \textit{Exact Match Accuracy (EMA)}, \textit{Final State Accuracy (FSA)}, and \textit{Step-wise Matching Score (SMS)}. 
\begin{itemize}
    \item \textbf{Exact Match Accuracy (EMA):} Measures whether the predicted action sequence exactly matches the ground truth:
    \begin{equation}
    \text{EMA} = \frac{1}{N} \sum_{i=1}^{N} \mathbb{I}[\hat{\mathcal{A}}_i = \mathcal{A}_i],
    \end{equation}
    where $N$ is the number of examples, $\hat{\mathcal{A}}_i$ is the predicted sequence, and $\mathcal{A}_i$ is the ground-truth sequence.

    \item \textbf{Final State Accuracy (FSA):} Assesses whether the final environment state resulting from $\hat{\mathcal{A}}_i$ matches the ground truth:
    \begin{equation}
    \text{FSA} = \frac{1}{N} \sum_{i=1}^{N} \mathbb{I}[\hat{\mathcal{S}}_i = \mathcal{S}_i],
    \end{equation}
    where $\hat{\mathcal{S}}_i$ is the predicted final state, and $\mathcal{S}_i$ is the ground-truth final state, \textit{i.e.}, the last frame of the demonstration video. In practice, we implement this by executing both the predicted and ground-truth programs in a lightweight simulation script to derive their final states, which avoids potential execution failures in a full simulator and enables reliable state comparison. 

    \item \textbf{Step-wise Matching Score (SMS):} Measures the fraction of the sequence correctly executed before deviation:
    \begin{equation}
    \text{SMS} = \frac{1}{N} \sum_{i=1}^{N} \frac{l_i}{T_i},
    \end{equation}
    where $T_i$ is the length of $\mathcal{A}_i$, and $l_i$ is the length of the matching prefix between $\hat{\mathcal{A}}_i$ and $\mathcal{A}_i$. 
    We adopt prefix-based evaluation over subsequence matching to account for the step-wise dependencies in long-horizon tasks. Since later actions often rely on the successful execution of earlier ones, a single early error can invalidate the remainder. Thus, prefix length better reflects partial execution success than unordered or sparse matches.
\end{itemize}

\subsection{Dataset Collection}
To ensure that each demonstration in our dataset is both semantically meaningful and physically feasible, the collection of tasks undergoes the following three stages: (1) task plan generation, (2) simulation-based validation, and (3) real-world human demonstration recording. 

Specifically, for each task category and difficulty level, we design prompt templates to 
guide GPT-4o in generating diverse task plans, which are manually reviewed for semantic and logical correctness. Validated plans are tested in a PyBullet simulation with a UR5e robot to ensure physical feasibility.
Infeasible plans are revised or discarded. Finally, each validated plan is enacted by a human demonstrator in a real-world tabletop setup, and recorded twice under two distinct environmental conditions. The first condition features a controlled setup with fixed lighting, camera position, and a clean background, providing a standardized evaluation environment. The second condition introduces variations in lighting and camera viewpoints to simulate more realistic, unconstrained scenarios. 
In total, LongVILBench comprises 150 tabletop manipulation tasks and 300 RGB videos, with a total of 2,332 annotated actions (7.8 actions per video on average) and six spatial relation types. The object set includes 14 items. Please see supplementary materials for more detailed statistics of the LongVILBench.

\begin{table*}[t]
\centering
\small
\begin{tabular}{lcccccccccccc}
\toprule
\multirow{2}{*}{\textbf{Method}} 
& \multicolumn{3}{c}{\textbf{Level1}} 
& \multicolumn{3}{c}{\textbf{Level2}} 
& \multicolumn{3}{c}{\textbf{Level3}} 
& \multicolumn{3}{c}{\textbf{Total}} \\
\cmidrule(lr){2-4} \cmidrule(lr){5-7} \cmidrule(lr){8-10} \cmidrule(lr){11-13}
 & EMA & FSA & SMS & EMA & FSA & SMS & EMA & FSA & SMS & EMA & FSA & SMS \\
\midrule
GPTforRobots-GPT4o   & 0.34 & 0.34 & 0.8050 & 0.14 & 0.14 & 0.4803 & 0.15 & 0.15 & 0.4324 & 0.21  & 0.21  & 0.5760 \\
SeeDo-GPT4o          & 0.35 & 0.35 & 0.5425 & 0.06 & 0.06 & 0.2198 & 0.00 & 0.00 & 0.1132 & 0.1367  & 0.1367  & 0.2919 \\
\midrule
Ours-Base-Qwen       & 0.68 & 0.68 & 0.8185 & 0.28 & 0.28 & 0.5660 & 0.12 & 0.12 & 0.4649 & 0.36  & 0.36  & 0.6063 \\                  
Ours-Reflection-Qwen & 0.73 & 0.73 & 0.8660 & 0.30 & 0.30 & 0.5380 & 0.17 & 0.17 & \textbf{0.4904} & 0.40  & 0.40  & 0.6348 \\
Ours-Base-GPT4o      & 0.76 & 0.76 & 0.8700 & 0.34 & 0.34 & 0.5635 & 0.19 & 0.19 & 0.4217 & 0.43  & 0.43  & 0.6184 \\
\textbf{Ours-Reflection-GPT4o}  
                     & \textbf{0.81} & \textbf{0.81} & \textbf{0.9000}  
                     & \textbf{0.40} & \textbf{0.40} & \textbf{0.6183}  
                     & \textbf{0.25} & \textbf{0.26} & 0.4649  
                     & \textbf{0.4867} & \textbf{0.49} & \textbf{0.6611} \\
\bottomrule
\end{tabular}
\caption{Experimental results on LongVILBench across three task levels. Best results are highlighted in \textbf{bold}.}
\label{tab:main_results}
\end{table*}
\section{Experiments}

We evaluate the proposed LongVIL framework on LongVILBench. Our evaluation includes comparisons against two representative video-to-code baselines: SeeDo~\cite{seedo} and GPT-4V for Robots~\cite{wake2024gpt} (abbreviated as \textbf{GPTforRobots}), two state-of-the-art VLM: GPT-4o~\cite{achiam2023gpt} and Qwen-VL-Max~\cite{bai2023qwenvlversatilevisionlanguagemodel} (abbreviated as \textbf{Qwen}).  
Additionally, we assess the framework's real-world applicability using a physical UR5e robotic arm.

\begin{figure}[h]
    \centering
    \includegraphics[width=0.9\linewidth]{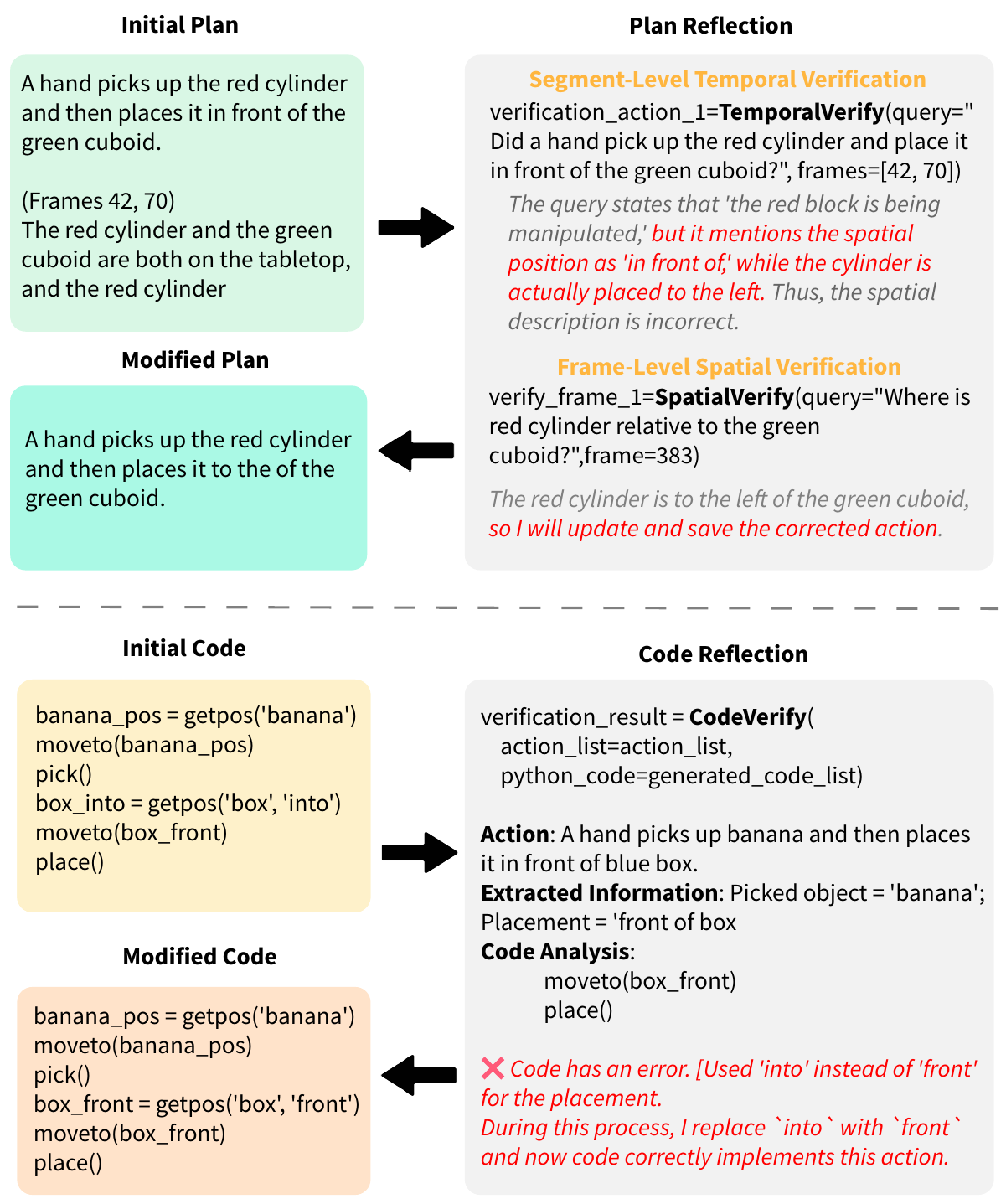}
    \caption{Qualitative comparison between the baseline agent and the agent with reflection modules. }
    \label{fig:qualitative}
\end{figure}

\subsection{Experimental Setup}

\subsubsection{Baselines}
We compare with two representative video-to-code baselines: \textbf{SeeDo} and \textbf{GPTforRobots}. To ensure a fair comparison, both baselines are implemented using the GPT-4o vision-language model (VLM) as their backbone. 
We modified the system prompts of these two methods to align their temporal and spatial relations with ours. 

\subsubsection{Implementation Details} 
We use `Base' to denote our framework with only the plan generation and code generation module, and use `Reflection' to denote our framework with two reflection modules. We use Qwen-VL-Max or GPT-4o as the VLM in our tools. 



\subsection{Quantitative Results}
Table~\ref{tab:main_results} summarizes the performance of all methods on LongVILBench across three difficulty levels. We highlight the following observations:

\begin{table*}[t]
\centering
\small
\begin{tabular}{lcccccccccccc}
\toprule
\multirow{2}{*}{\textbf{Configuration}} 
& \multicolumn{3}{c}{\textbf{Level1}} 
& \multicolumn{3}{c}{\textbf{Level2}} 
& \multicolumn{3}{c}{\textbf{Level3}} 
& \multicolumn{3}{c}{\textbf{Total}} \\
\cmidrule(lr){2-4} \cmidrule(lr){5-7} \cmidrule(lr){8-10} \cmidrule(lr){11-13}
 & EMA & FSA & SMS & EMA & FSA & SMS & EMA & FSA & SMS & EMA & FSA & SMS \\
\midrule
A (Base) &
0.76 & 0.76 & 0.8700 & 
0.34 & 0.34 & 0.5635 & 
0.19 & 0.19 & 0.4217 & 
0.43 & 0.43 & 0.6184 \\
B (+Keyframe Completion) & 
0.78 & 0.79 & 0.8926 & 
0.35 & 0.35 & 0.5903 & 
0.19 & 0.21 & 0.4500 & 
0.44 & 0.45 & 0.6439 \\
C (+$\mathcal{R}_{plan}$, Plan Reflection) & 
0.80 & 0.80 & 0.8975 & 
0.37 & 0.37 & 0.5718 & 
0.23 & 0.24 & \textbf{0.4766} & 
0.4667 & 0.47 & 0.6486 \\
D (+$\mathcal{R}_{code}$, Code Reflection) & 
\textbf{0.81} & \textbf{0.81} & \textbf{0.9000} & 
\textbf{0.40} & \textbf{0.40} & \textbf{0.6183} & 
\textbf{0.25} & \textbf{0.26} & 0.4649 & 
\textbf{0.4867} & \textbf{0.49} & \textbf{0.6611} \\
\bottomrule
\end{tabular}
\caption{
Ablation results on LongVILBench across three task levels. Best results are highlighted in bold. 
}
\label{tab:ablation}
\end{table*}

\begin{table}[t]
\centering
\small
\setlength{\tabcolsep}{4pt}
\scalebox{0.9}{
\begin{tabular}{l@{\hskip 6pt}ccc@{\hskip 8pt}ccc}
\toprule
\multirow{2}{*}{\textbf{Method}} 
& \multicolumn{3}{c}{\textbf{Clean}} 
& \multicolumn{3}{c}{\textbf{Complex}} \\
\cmidrule(lr){2-4} \cmidrule(lr){5-7}
 & EMA & FSA & SMS & EMA & FSA & SMS \\
\midrule
GPTforRobots         & 0.22 & 0.22 & 0.59 & 0.20 & 0.20 & 0.56 \\
SeeDo                & 0.14 & 0.14 & 0.29 & 0.13 & 0.13 & 0.29 \\
Ours-Base-Qwen       & 0.41 & 0.41 & 0.63 & 0.32 & 0.32 & 0.58 \\
Ours-Reflection-Qwen & 0.44 & 0.44 & 0.66 & 0.36 & 0.36 & 0.61 \\
Ours-Base-GPT4o      & 0.47 & 0.47 & 0.65 & 0.39 & 0.39 & 0.58 \\
\textbf{Ours-Reflection-GPT4o}  
                     & \textbf{0.55} & \textbf{0.56} & \textbf{0.71}
                     & \textbf{0.42} & \textbf{0.42} & \textbf{0.61} \\
\bottomrule
\end{tabular}
}
\caption{Performance under the clean and complex visual conditions in LongVILBench. Hyperparameter $\Delta$ in keyframe completion in is set to 100 in clean visual conditions, and 50 in complex visual conditions.}
\label{tab:vis_cond}
\end{table}

\subsubsection{Significant improvements over baselines} 
Our best configuration (\textbf{Ours-Reflection-GPT4o}) achieves substantial gains in all metrics compared to GPTforRobots and SeeDo. For example, the overall EMA (0.21~$\rightarrow$~0.4867, +131.8\%) and FSA (0.21~$\rightarrow$~0.49, +133.3\%) are both more than double those of GPTforRobots, and SMS (0.5760~$\rightarrow$~0.6611, +14.8\%) also shows a notable improvement. These results confirm that the proposed reflection framework effectively addresses the compounding errors.

\subsubsection{Reflection consistently boosts performance} 
Within both GPT-4o and Qwen-VL-Max, the reflection-enhanced variant outperforms the Base configuration across all metrics. For GPT-4o, reflection yields improvements in EMA (0.43~$\rightarrow$~0.4867, +13.2\%), FSA (0.43~$\rightarrow$~0.49, +14.0\%), and SMS (0.6184~$\rightarrow$~0.6611, +6.9\%), demonstrating that the plan and code reflection modules jointly refine action plans for better sequence fidelity and final-state correctness, no matter what large models are used.

\subsubsection{Robustness on long-horizon tasks} 
All methods experience performance degradation from Level~1 to Level~3 due to increased sequence length and complexity. However, \textbf{Ours-Reflection-GPT4o} shows the smallest drop, maintaining substantially higher EMA (0.15~$\rightarrow$~0.25, +66.7\%) and SMS (0.4324~$\rightarrow$~0.4649, +7.5\%) on Level~3 compared to GPTforRobots, and much larger gains over SeeDo (0.00~$\rightarrow$~0.25, +0.25 absolute EMA; 0.1132~$\rightarrow$~0.4649, +310\% SMS), indicating better resilience to long-horizon challenges.

\subsection{Qualitative Results}
Figure~\ref{fig:qualitative} illustrates a representative example highlighting the effectiveness of our two reflection modules. The visual and code reflection modules collaboratively detect and correct errors in the initial plan and generated code, respectively.

\subsection{Ablation Study}
We perform an ablation study to evaluate the contribution of each module in our framework. Table~\ref{tab:ablation} presents four configurations: (A) base pipeline with keyframe-based planning, (B) adding temporal augmentation (+Keyframe Completion), (C) further incorporating plan reflection module (\(\mathcal{R}_{plan}\)), and (D) further incorporating code reflection module (\(\mathcal{R}_{code}\)).

We highlight the following observations: (1) Temporal augmentation improves the completeness of action representation by interpolating intermediate keyframes, leading to notable gains, \textit{e.g.}, Level~1 SMS (0.8700~$\rightarrow$~0.8926, +2.6\%) and Level~3 FSA (0.19~$\rightarrow$~0.21, +10.5\%), thereby mitigating action omissions. (2) Incorporating the video-based reflection module \(\mathcal{R}_{plan}\) further enhances spatiotemporal alignment, improving overall EMA (0.44~$\rightarrow$~0.4667, +6.1\%) and increasing Level~3 SMS (0.4500~$\rightarrow$~0.4766, +5.9\%), underscoring the role of visual feedback in verifying temporal consistency and spatial relations. (3) Finally, adding the code-level reflection module \(\mathcal{R}_{code}\) ensures semantic coherence between the high-level plan and executable code, yielding the highest overall performance (EMA: 0.4867, FSA: 0.49, SMS: 0.6611). Our framework shows the greatest improvements on Level~3 tasks (EMA: 0.19~$\rightarrow$~0.25, +31.6\%; FSA: 0.19~$\rightarrow$~0.26, +36.8\% over baseline), demonstrating that the importance of the two reflection modules for long-horizon visual imitation.

\subsection{Results under Varying Visual Conditions}
Table~\ref{tab:vis_cond} reports results under two visual conditions: `clean' (fixed lighting, static camera, clean background) and `complex' (variable lighting, dynamic viewpoints). All methods exhibit performance drops under complex settings, highlighting the challenges posed by real-world visual variability. These findings validate LongVILBench as a robust benchmark for evaluating visual imitation under diverse conditions.

\begin{figure}[h]
    \centering
    \includegraphics[width=\linewidth]{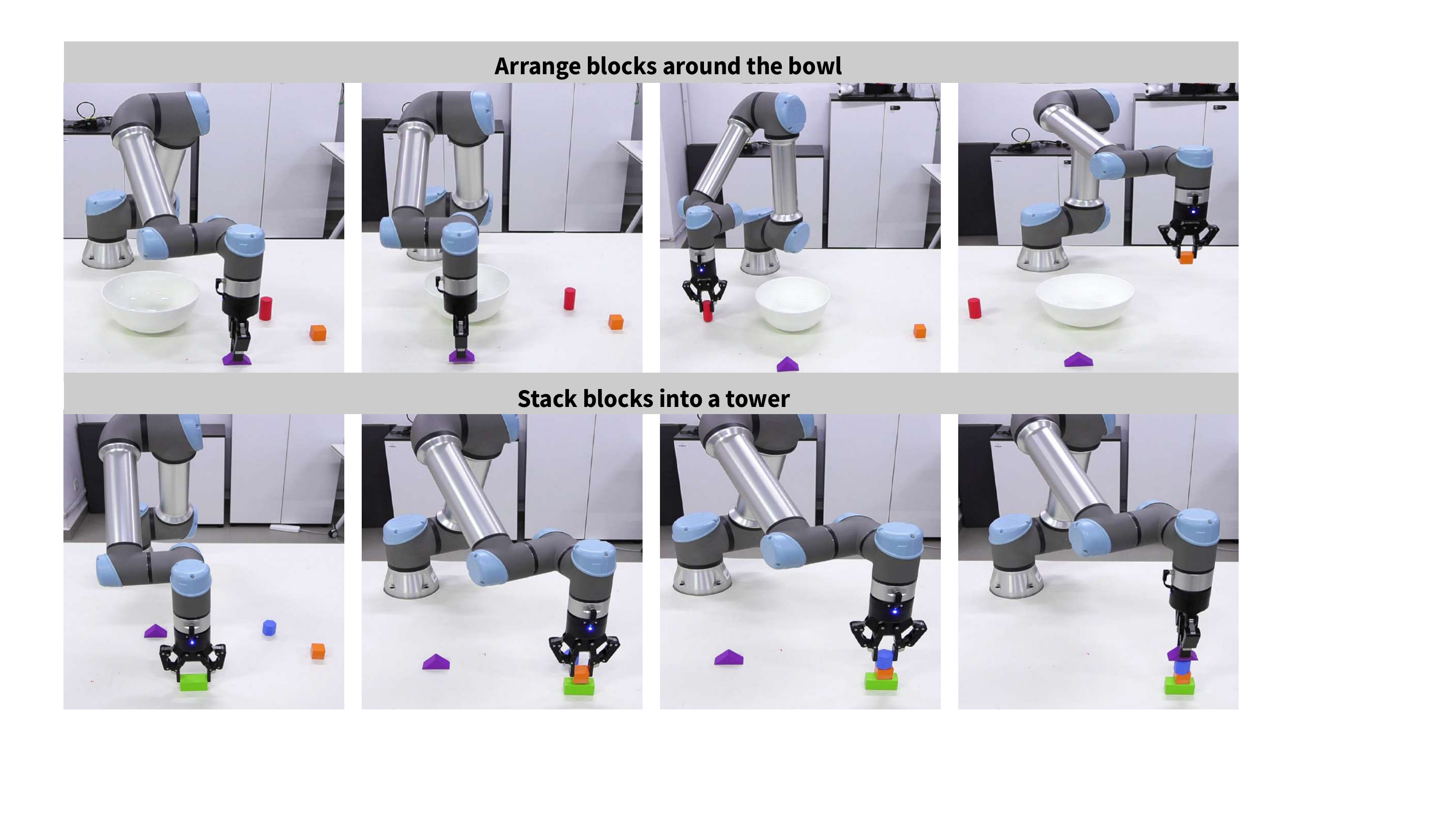}
    \caption{Real-world execution of symbolic plans on the UR5e robot. These examples demonstrate successful transfer of the generated programs from simulation to real-world settings.}
    \label{fig:realworld}
\end{figure}

\subsection{Real-World Deployment}

To evaluate real-world applicability, we deploy the generated symbolic programs $\Pi^*$ on a UR5e robot arm~\cite{ur5e} with a Robotiq 2F-85 gripper~\cite{robotiq2f85} and an overhead RGB camera~\cite{hikvisioncam}. The system uses the urx library~\cite{urx} for control and a Molmo-based model~\cite{molmo} for object detection and pose estimation. All high-level actions are translated into motion primitives, including grasping, placement, and stacking with height-aware control.

We test the full system on 20 subset tasks from LongVILBench, spanning all difficulty levels, and achieve an 85\% success rate (17/20). Most failures stem from hardware issues (\textit{e.g.}, gripper misalignment), not from planning or execution logic. These results confirm that our framework generalizes from simulation to physical deployment with minimal adaptation and no additional training. Figure~\ref{fig:realworld} shows representative task executions involving object arrangement and stacking.

\section{Discussions and Conclusions}



We presented LongVIL, a novel framework for long-horizon visual imitation learning that integrates two reflection modules at both the plan and code levels. By introducing two dedicated reflection modules—plan reflection for verifying temporal and spatial alignment with demonstration videos, and code reflection for ensuring semantic consistency between actions and executable programs—our framework significantly improves robustness and accuracy in complex, multi-step tasks. To support systematic evaluation, we introduced LongVILBench, a new benchmark comprising 300 real-world human demonstrations across diverse manipulation domains and varying difficulty levels. Extensive experiments demonstrate that LongVIL outperforms existing video-to-code baselines, particularly under long-horizon and visually complex conditions. 
Overall, our results highlight the value of structured self-verification in scaling visual imitation learning to long-horizon, real-world scenarios.








\bibliographystyle{IEEEtranBST/IEEEtran} 
\bibliography{IEEEtranBST/ref}

\end{document}